\documentclass[11pt,a4paper]{article}
\usepackage[utf8]{inputenc}
\usepackage[hyperref]{acl2018}
\usepackage{mathptmx}
\usepackage{latexsym}

\usepackage{url}

\usepackage{graphicx}
\usepackage{amssymb}
\usepackage{amsmath}
\usepackage{epstopdf}
\usepackage{subcaption}
\usepackage{xcolor}
\usepackage{metalogo}
\usepackage{tikz}
\usepackage{enumitem}

\aclfinalcopy 
\def\aclpaperid{***} 

\usetikzlibrary{arrows, decorations.markings, decorations.pathmorphing, positioning}
\usetikzlibrary{fit}					
\usetikzlibrary{backgrounds}	
\usepackage{pgfgantt}
\usetikzlibrary{mindmap,backgrounds,trees}
\usetikzlibrary{positioning,shapes,shadows,arrows}

\tikzstyle{state}=[rectangle,rounded corners,
                                    thick,
                                    minimum size=1.2cm,
                                    draw=blue!80,
                                    fill=blue!20]

\tikzstyle{input}=[rectangle,rounded corners,
                                                thick,
                                                minimum size=1.2cm,
                                                draw=orange!80,
                                                fill=orange!25]

\tikzstyle{output}=[rectangle,rounded corners,
                                    thick,
                                    minimum size=1.2cm,
                                    draw=purple!80,
                                    fill=purple!20]

\tikzstyle{matrx}=[rectangle,
                                    thick,
                                    minimum size=1cm,
                                    draw=gray!80,
                                    fill=gray!20]

\tikzstyle{noise}=[circle,
                                    thick,
                                    minimum size=1.2cm,
                                    draw=yellow!85!black,
                                    fill=yellow!40,
                                    decorate,
                                    decoration={random steps,
                                                            segment length=2pt,
                                                            amplitude=2pt}]

\tikzstyle{background}=[rectangle,
                                                fill=gray!10,
                                                inner sep=0.2cm,
                                                rounded corners=5mm]

\title{What is not where: the challenge of integrating spatial representations into deep learning architectures}

\author{John D. Kelleher\\ 
ADAPT Centre for Digital Content Technology\\
Dublin Institute of Technology, Ireland\\
\texttt{john.d.kelleher@dit.ie} \\\And
Simon Dobnik\\
CLASP and FLOV\\
University of Gotenburg, Sweden\\
\texttt{simon.dobnik@gu.se}}

\date{}

\begin{document}
\maketitle
\begin{abstract}
This paper examines to what degree current deep learning architectures for image caption generation capture spatial language. On the basis of the evaluation of examples of generated captions from the literature we argue that systems capture \emph{what} objects are in the image data but not \emph{where} these objects are located: the captions generated by these systems are the output of a language model conditioned on the output of an object detector that cannot capture fine-grained location information. Although language models provide useful knowledge for image captions, we argue that deep learning image captioning architectures should also model geometric relations between objects. 
\end{abstract}

\section{Introduction}\label{sec:introduction}
There is a long-traditional in Artificial Intelligence (AI) of developing computational models that integrate language with visual information, see \emph{inter alia.}: \cite{winograd73:_comput_model_thoug_languag,mckevitt9596:NLV,kelleher:2003,gorniak/roy:2004,kelleher2005contextNLG,brenner2007mediating,Dobnik:2009dz,tellex2010natural,Sjoo:2011aa,dobnik2016model,schutte2017robot}. 
The goal of this paper is to situate and critically examine recent advances in computational models that integrate visual and linguistic information.
One of the most exciting developments in AI in recent years has been the development of \emph{deep learning} (DL) architectures \cite{lecun2015deep,Schmidhuber201585}. Deep learning models are neural network models that have multiple hidden layers. The advantage of these architectures is that these models have the potential to learn high-level useful features from raw data. For example, \citet{Lee:2009:CDB:1553374.1553453} report how their \emph{convolutional deep belief network} ``learns useful high-level visual features, such as object parts, from unlabelled images of objects and natural scenes''. In brief, \citeauthor{Lee:2009:CDB:1553374.1553453} show how a deep network trained to perform face recognition learns a hierarchical sequence of feature abstractions: neurons in the early layers in the network learn to act as edge detectors, neurons in later layers react to the presence of meaningful parts of a face (e.g., nose, eye, etc.), and the neurons in the last layers of the network react to sensible configurations of body parts (e.g., nose and eyes and sensible (approximate) offsets between them).

\begin{figure*}[htbp]
\includegraphics[width=\textwidth]{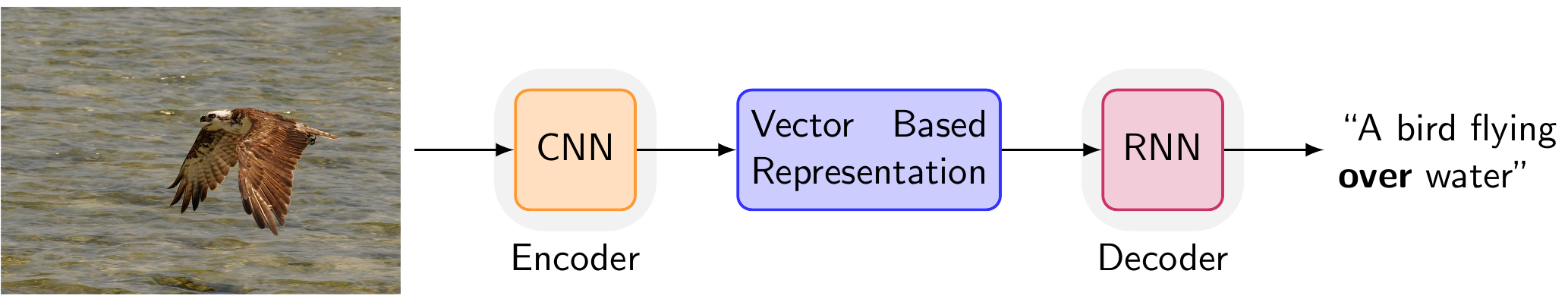} 
\caption{A schematic of a typical encoder-decoder deep learning image captioning architecture 
in \cite{xu2015show}. The photo is different from but similar to the photo in this example. The current photo is by Jerry Kirkhart (originally posted to Flickr as Osprey Hunting) and is sourced via Wikimedia Commons. It is used here under the Creative Commons Attribution 2.0 generic licence.}
\label{fig:imagecaptioningbird}
\end{figure*}

Deep learning models have improved on the start-of-the-art across a range of image and language modelling tasks. The typical deep learning architecture for image modelling is a \emph{convolutional neural network} (CNN) \cite{726791} and for language modelling is a \emph{recurrent neural network} (RNN), often using \emph{long short-term memory} (LSTM) units  \cite{doi:10.1162/neco.1997.9.8.1735}. However, from the perspective of research into the interface between language and vision perhaps the most exciting aspect of deep learning is the fact that all of these models (both language and vision processing architectures) use a vector based representation. A consequence of this is that deep learning models have the potential to learn multi-modal representations that integrate linguistic and visual information. Indeed, inspired by sequence-to-sequence neural machine translation research \cite{sutskever2014sequence}, deep learning image captioning systems have been developed that use a CNN to process and encode image data and then pass this vector based encoding of the image to an RNN that generates a caption for the image. Figure~\ref{fig:imagecaptioningbird} illustrates the components and flow of data in an encoder-decoder CNN-RNN image captioning architecture.   

 
The performance of these deep learning image captioning systems is impressive. However, the question posed by this paper is whether these systems are actually grounding the semantics of the entire linguistic caption in the image, and in particular whether these systems ground the semantics of spatial relations in the image. The rest of the paper is structured as follows: Section~\ref{sec:captioning} introduces the components of deep learning image captioning architecture in more detail; following this, Section \ref{sec:grounding} reviews the challenges of grounding language in perception, with a particular focus on spatial language; the paper concludes in Section \ref{sec:conclusions} by posing the question of whether deep learning image captioning architectures as currently constituted are capable of doing justice to the complexity and diversity of factors that affect the production and interpretation of spatial language in visually situated dialogues.


\section{The Standard DL Image Captioning Architecture}
\label{sec:captioning}

As mentioned in Section~\ref{sec:introduction} there are two major components within current standard deep-learning image captioning systems, a CNN that processes the image input and encodes some of the information from the image as a vector, and an RNN that takes the vector representation of the image as an input and generates a caption for the image. This section provides an explanation for how each of these components works: Section~\ref{sec:cnn} introduces the basic architecture of a CNN and Section~\ref{sec:rnn} introduces the basic architecture of an RNN and explains how they can be used to create visually grounded language models.


\subsection{Convolutional Neural Networks}
\label{sec:cnn}

CNNs are specifically designed for image recognition tasks, such as handwritten digit recognition \cite{lecun:1989}. A well-recognised approach to image recognition is to extract local visual features and combine these features to form higher-order features. A local feature is a feature whose extent within a image is constrained to a small set of neighbouring pixels. For example, for face recognition a system might first learn to identify features such as patches of line or curve segments, and then learn patterns across these low level features that correspond to features such as eyes or a mouth, and finally learn how to combine these body-part features to identify a face.

A key challenge in image recognition is creating a model that is able to recognise if a visual feature has occurred in the image irrespective of the location of the feature in the image:

\begin{quote}
``it seems useful to have a set of feature detectors that can detect a particular instance of a feature anywhere on the input plane. Since the precise location of a feature is not relevant to the classification, we can afford to loose some position information in the process'' \cite[p.14]{lecun:1989}
\end{quote}

\noindent For example, a face recognition network should recognise the shape of an eye whether the eye is in the top right corner of the image or in the centre of the image. CNNs achieve this translation invariant detection of local visual features using two techniques:
\begin{enumerate}[noitemsep]
	\item weight (parameter) sharing and
	\item pooling.
\end{enumerate}

\noindent Recall that each neuron in a network learns a function that maps from a set of inputs to an output activation. The function is defined by the set of weights the neuron applies to the inputs it receives and learning the function involves updating the weights from a set of random initialised values to a set of values that define a function that the network found useful during training in terms of predicting the correct output value. In the context of image recognition a function can be understood as a feature detector which takes a set of pixel values as input and outputs a high-activation score if the visual feature is present in the set of input pixels and a low-activation score if the feature is not present. Furthermore, neurons that share (or use) the same weights implement the same function and hence implement that same feature detector. 


Given that a set of weights for a neuron defines a feature detector and that neurons with the same weights implement the same feature detector, it is possible to design a network to check whether a visual feature occurs anywhere in an image by making multiple neurons share the same set of weights but have each of these neurons inspect different portions of the image in such a way so that together the neurons cover the whole image. 

\begin{figure}[htbp]
        \includegraphics[width=\columnwidth]{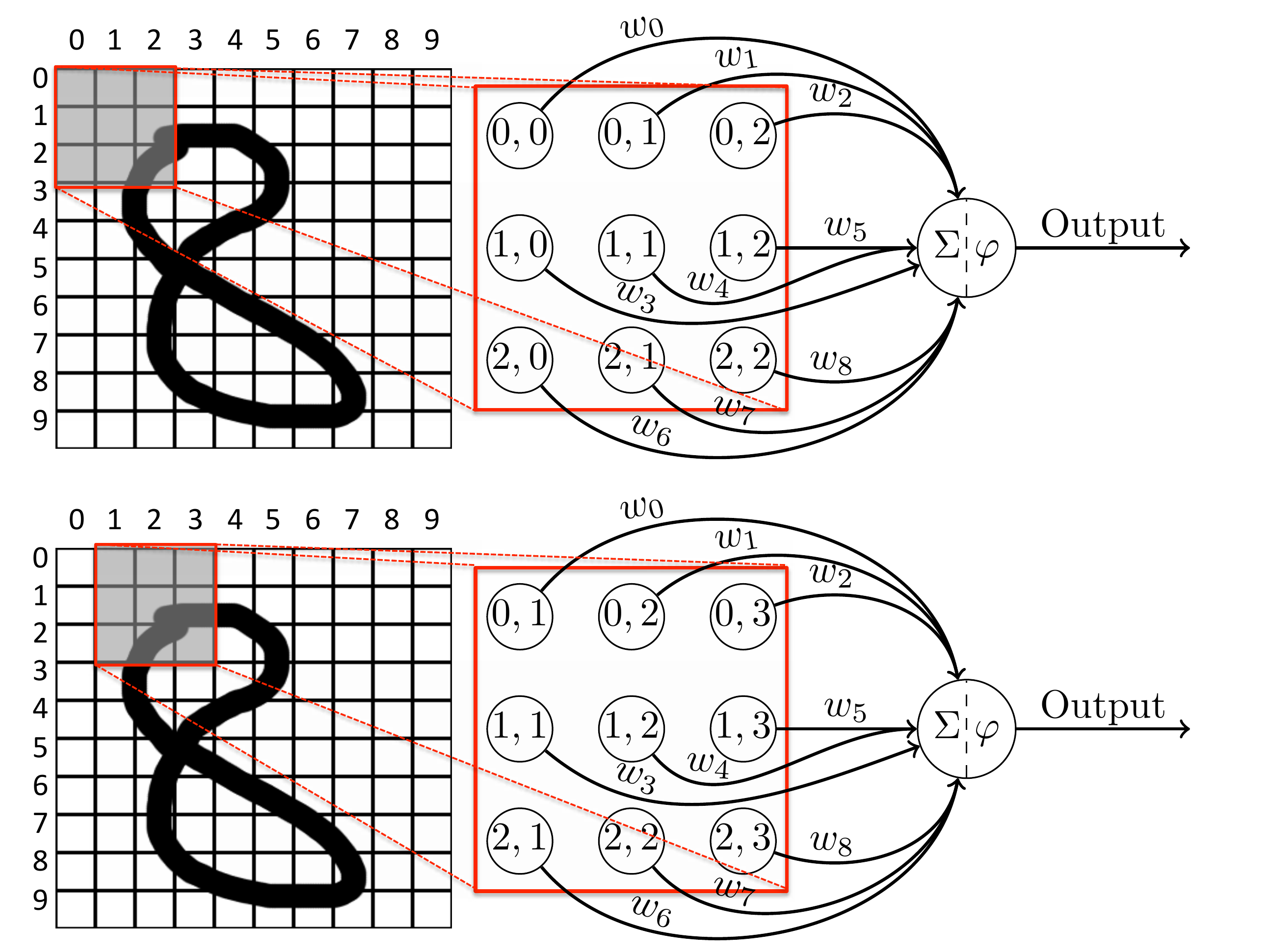} 
\caption{Two neurons connected to different areas in the input image (i.e., with different receptive fields) but which shared the same weights: $w_0, w_1, w_2, w_3, w_4, w_5, w_6, w_7, w_8$}
\label{fig:cnn-parametersharing}
\end{figure}

For example, imagine we wish to train a network to identify digits in images of $10\times 10$ pixels. In this scenario we may design a network so that one of the neurons in the network inspects the pixels in the top-left corner of the figure to check if a visual feature is present. The image at the top of Figure~\ref{fig:cnn-parametersharing} illustrates such a neuron. This neuron inspects the pixels $(0,0), \dots ,(2,2)$ and applies the function defined by the weight vector $<w_0, \dots, w_8>$. This neuron will return a high activation if the appropriate pixel pattern is present in the pixels input to the function and low otherwise. We can now create a copy of this neuron that uses the same weights $<w_0, \dots, w_8>$ but which inspects a different set of pixels in the image: the image at the bottom of Figure~\ref{fig:cnn-parametersharing} illustrates such a neuron, this particular neuron inspects the pixels $(0,1), \dots ,(2,3)$. If the visual feature that the function defined by the weight vector $<w_0, \dots, w_8>$ occurs in either of the image patches inspected by these two neurons, then one of the neurons will fire.

\begin{figure}[htbp]
\centerline{
        \includegraphics[width=0.7\columnwidth]{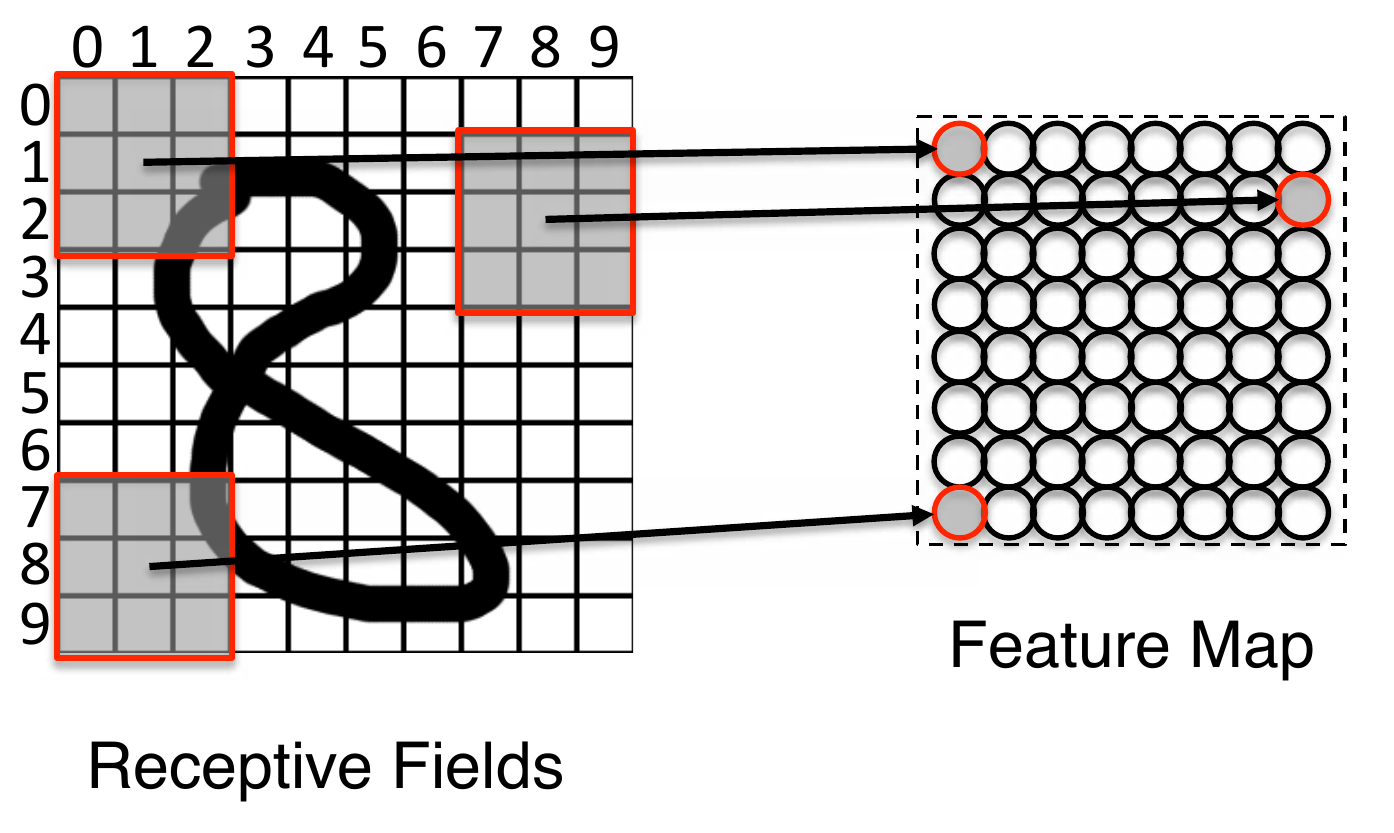} 
        }
\caption{A feature map }
\label{fig:cnn-featuremap}
\end{figure}

Extending the idea of a set of neurons with shared weights inspecting a full image results in the concept of a feature map. In a CNN a feature map consists of a group of neurons that share the same set of weights on their inputs. This means that each group of neurons that share their weights learns to identify a particular visual feature and each neuron in the group acts as a detector for that feature. In a CNN the neurons within each group are arranged so that each neuron examines a different local region in the image. The set of pixels that each neuron in the feature map inspects is known as the receptive field of that neuron. The neurons and the related receptive fields are arranged so that together the receptive fields cover the entire input image. Consequently, if the visual feature the group detects occurs anywhere in the image one of the neurons in the group will identify it.  Figure~\ref{fig:cnn-featuremap} illustrates a feature map and how each neuron in the feature map has a different receptive field and how the neurons and fields are organised so that taken together the receptive fields of the feature map cover the entire input image. Note that the receptive fields of neighbouring neurons typically overlap. In the architecture illustrated in Figure~\ref{fig:cnn-featuremap} the receptive fields of neighbouring neurons will overlap by either two columns (for horizontal neighbours) or by two rows (for vertical neighbours). However, an alternative organisation would be to reduce the number of neurons in the feature map and reduce the amount of overlap in the receptive fields. For example, if the receptive fields only overlapped by one row or one column then we would only need half the number of neurons in the feature map to cover the entire input image. This would of course result in a ``two-to-one under-sampling in each direction'' \cite{lecun:1989}.

\begin{figure}[htbp]
\centerline{
        \includegraphics[width=0.8\columnwidth]{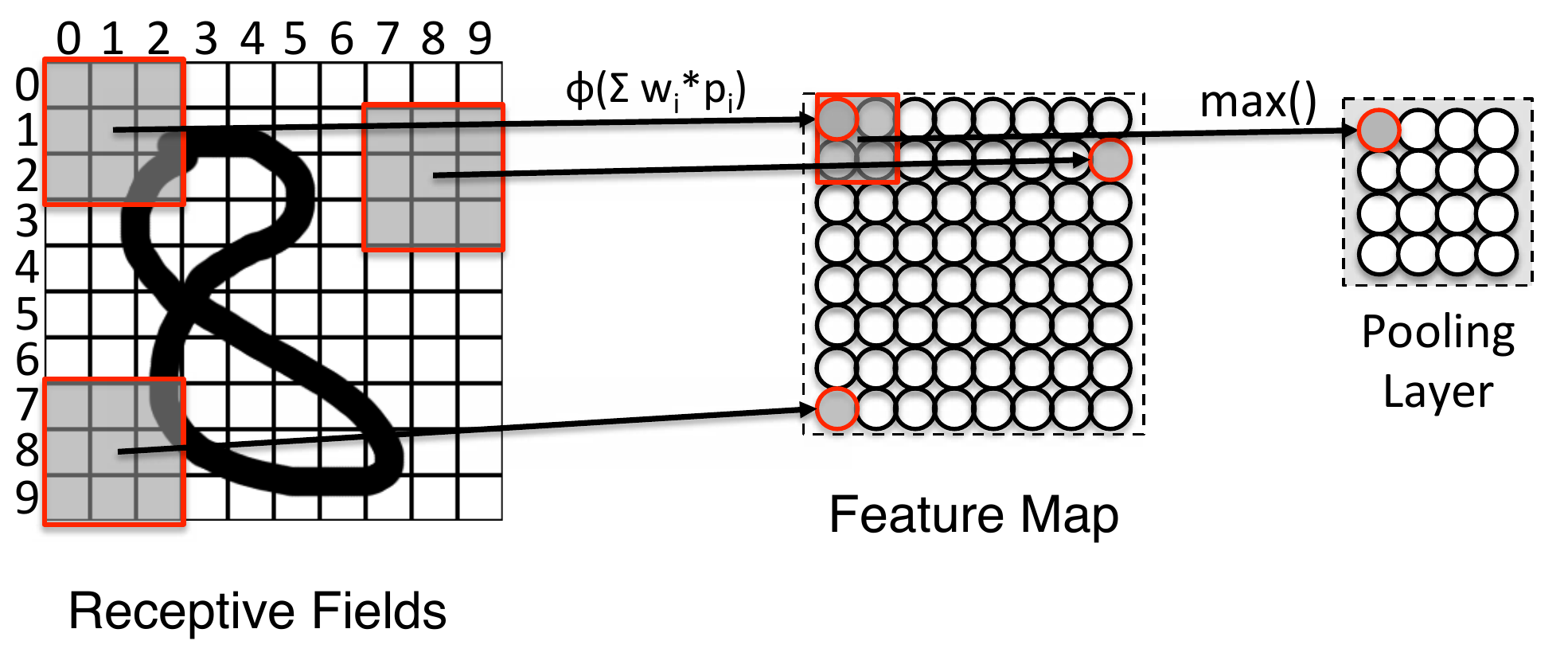} 
}
\caption{Applying pooling to a feature map}
\label{fig:cnn-featuremap-pooling}
\end{figure}

The idea of applying the same function repeatedly across an input space by defining a set of neurons where each neuron applies the function to a different part of the input is very general and can be used irrespective of the type of function being applied. CNNs networks often use the same technique to under-sample the output from a feature map. The motivation for under-sampling is to \emph{discard locational information in favour of generalising the network's ability to identify visual features in a shift invariant manner}. The standard way to implement under-sampling on the output of a feature map is to use a pooling layer, so called as it pools information from a number of neurons in a feature map. Each neuron in a pooling layer inspects the outputs of a subset of the neurons in a feature map, in a very similar way to the way the neuron in the feature map each has a receptive field in the input. Often the function used by neurons in a pooling layer is the \emph{max} function. Essentially, a max pooling neuron outputs the maximum activation value of any of the neurons in the preceding layer that it inspects. Figure~\ref{fig:cnn-featuremap-pooling} illustrates the extension of the feature map in Figure~\ref{fig:cnn-featuremap} with a pooling layer. The output of the highlighted neuron in the pooling layer is simply the highest activation across the 4 neurons in the feature map that it inspects. Pooling obviously discards locational information at a local level, after pooling the network knows that a visual feature occurred in a region of the image but does not know where precisely within the region the feature occurred.

 
\begin{figure}[htbp]
        \includegraphics[width=\columnwidth]{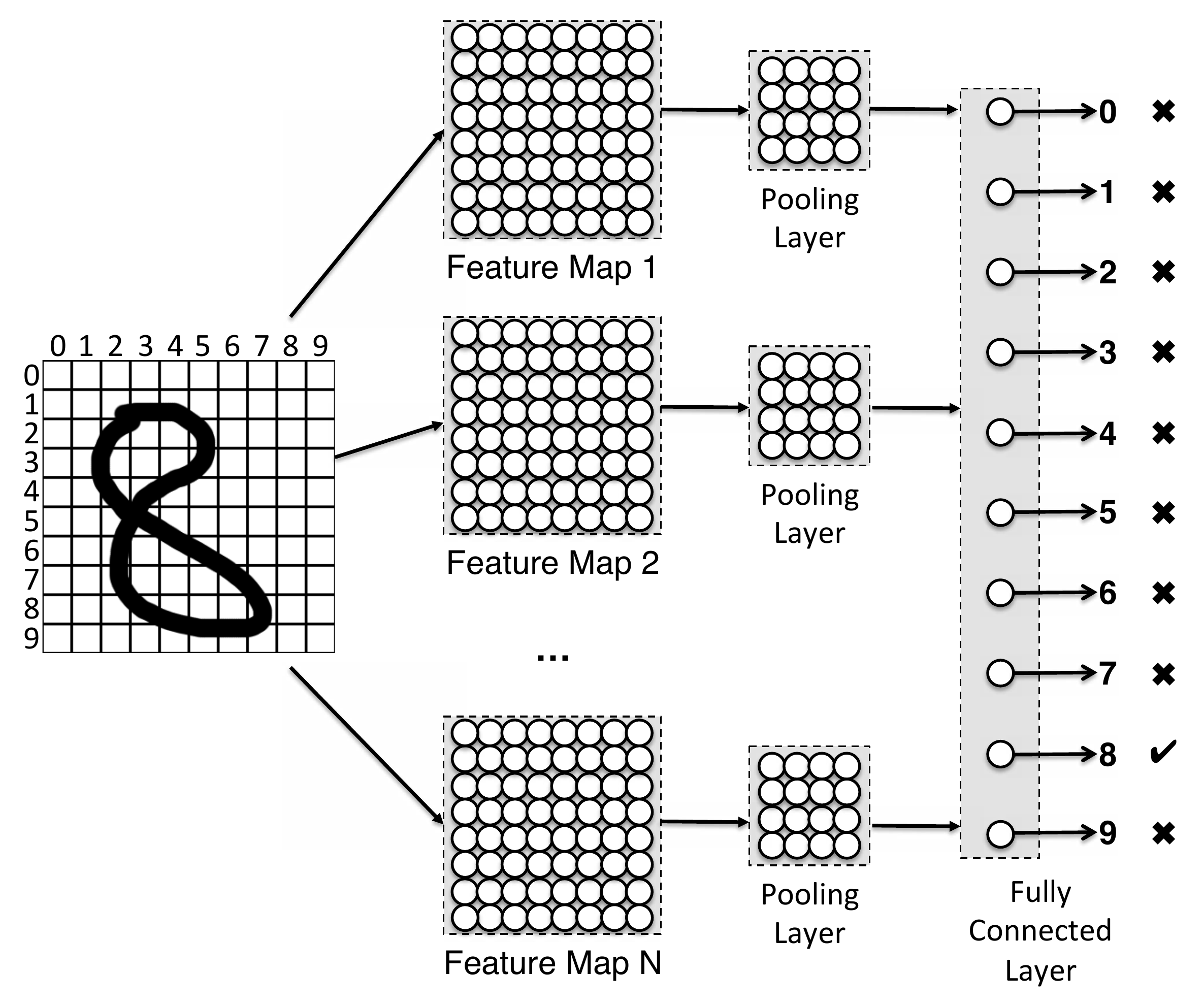} 
\caption{A CNN architecture containing \emph{N} parallel streams of feature maps and pooling layers feeding into a single fully connected feed-forward layer}
\label{fig:cnn-architecture}
\end{figure}

A CNN network is not restricted to only one feature map or one pooling layer. A CNN network can consist of multiple feature maps and pooling layers working in parallel where the outputs of these different streams of processing are finally merged to one or more fully connected layers (see Figure~\ref{fig:cnn-architecture}). Furthermore, these basic building blocks of feature maps and pooling layers can be sequenced in many different ways: the output of one feature map layer can be used as the input to another feature map layer, and the output of a pooling layer may be the input to a feature map layer. Consequently, a CNN architecture is very flexible and can be composed of multiple layers of feature maps and pooling layers.  For example, a CNN could include a feature map that is fed into a pooling layer which in turn acts as the input for a second feature map layer which itself is down-sampled using another pooling layer, and so on, until the outputs of a layer are eventually fed into a fully-connected feed-forward layer where the final prediction is calculated: feature map $\rightarrow$ pooling $\rightarrow$ feature map $\rightarrow$ pooling $\rightarrow$ $dots$ $\rightarrow$ fully-connected layer. Obviously with each extra layer of pooling the network discards more and more location information. 

\subsection{
Recurrent Neural Network Language Models}\label{sec:rnn}

\begin{figure*}[t!]
    \centering
    \begin{subfigure}[t]{0.5\textwidth}
        \centering
        \includegraphics[width=\textwidth]{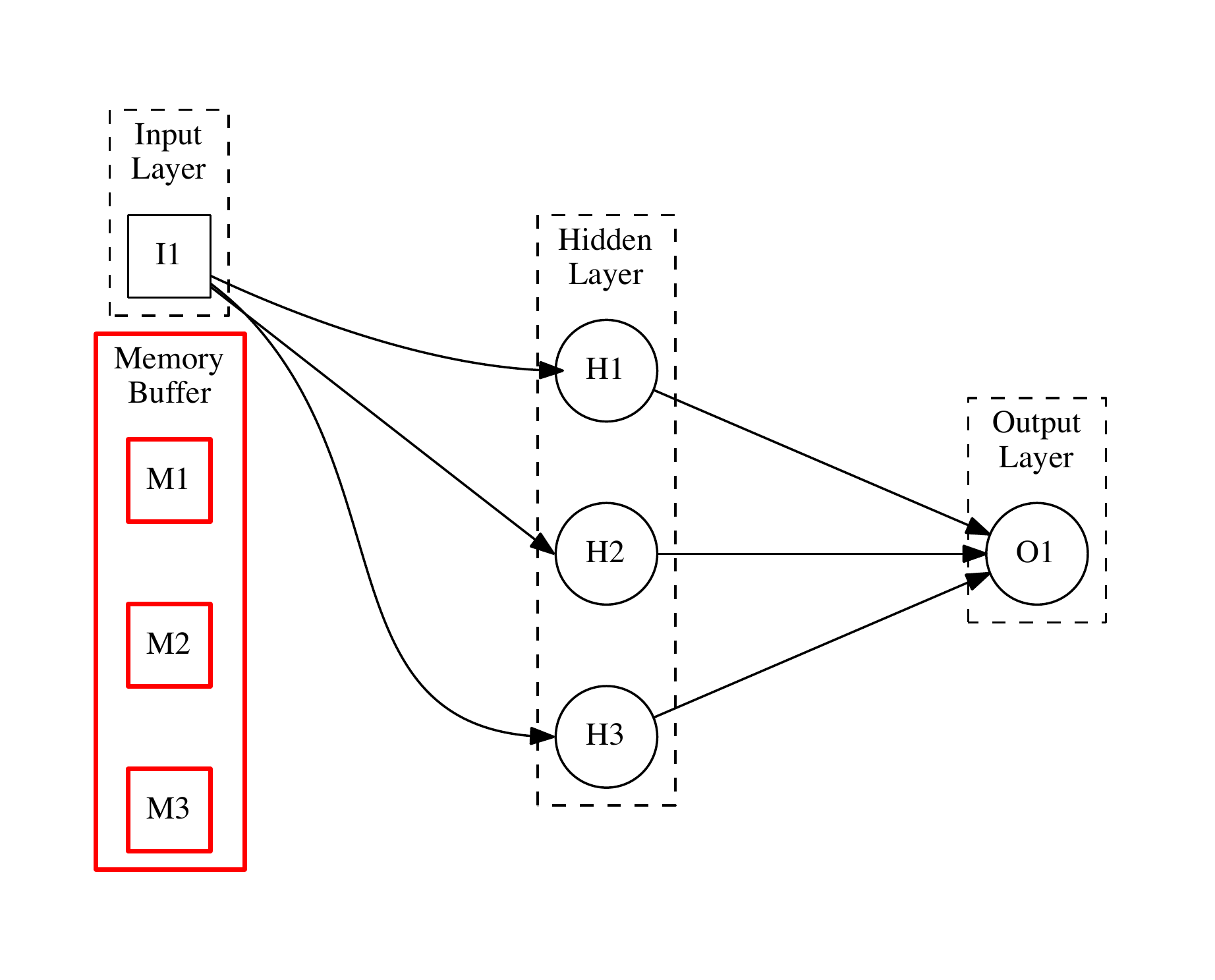}
\caption{Adding a memory buffer to a feed-forward neural network with one hidden layer}
\label{fig:rnn-memorybuffer}
    \end{subfigure}%
    ~
    \begin{subfigure}[t]{0.5\textwidth}
        \centering
        \includegraphics[width=\textwidth]{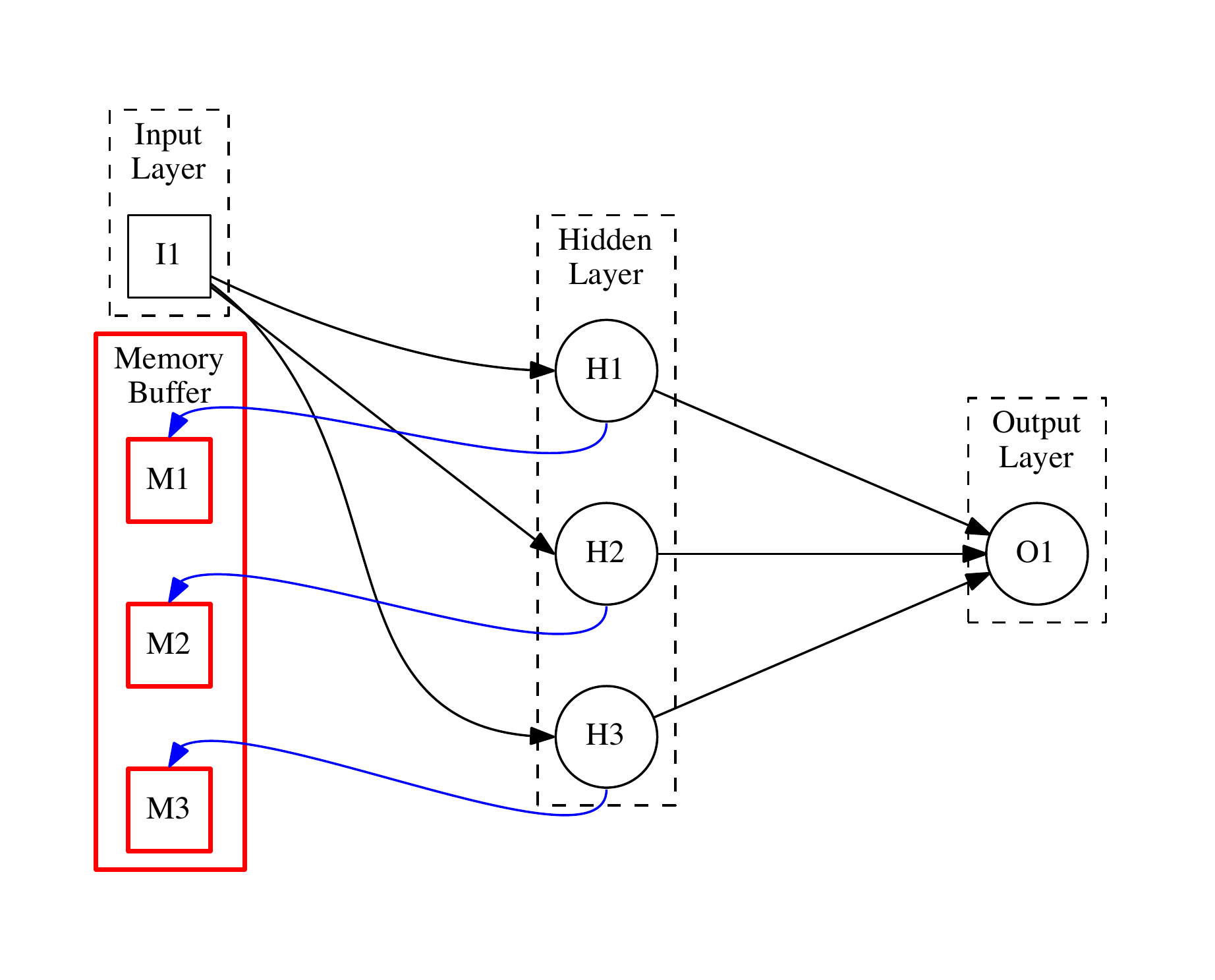}
\caption{Writing the activation of the hidden layer to the memory buffer after processing the input at time $t$}
\label{fig:rnn-writing-to-memory}
    \end{subfigure}
    \\
    \centering
    \begin{subfigure}[t]{0.5\textwidth}
        \centering
        \includegraphics[width=\textwidth]{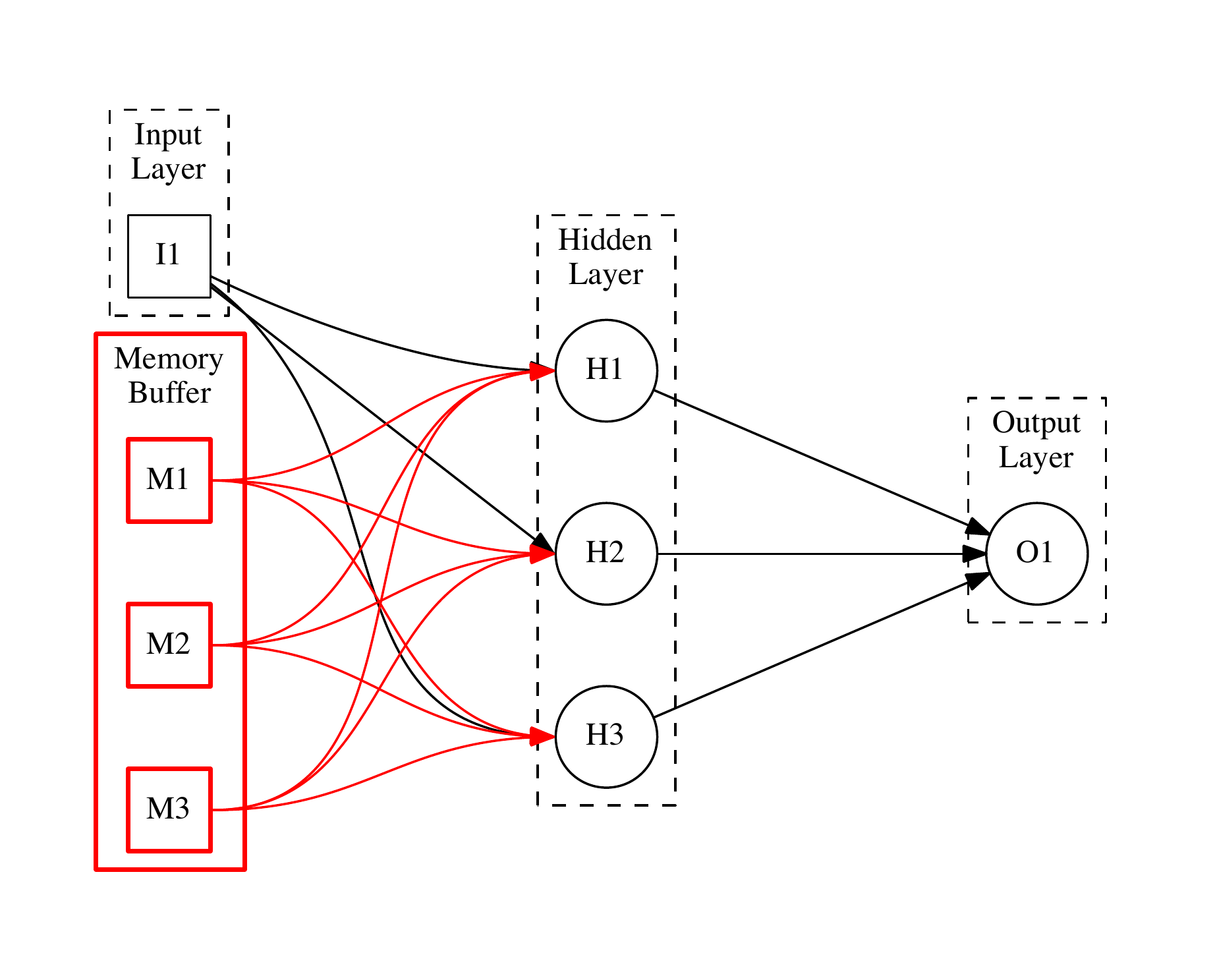}
\caption{Merging the memory buffer with the next input at $t+1$}
\label{fig:rnn-reading-from-memory}
    \end{subfigure}%
    ~
    \begin{subfigure}[t]{0.5\textwidth}
        \centering
        \includegraphics[width=\textwidth]{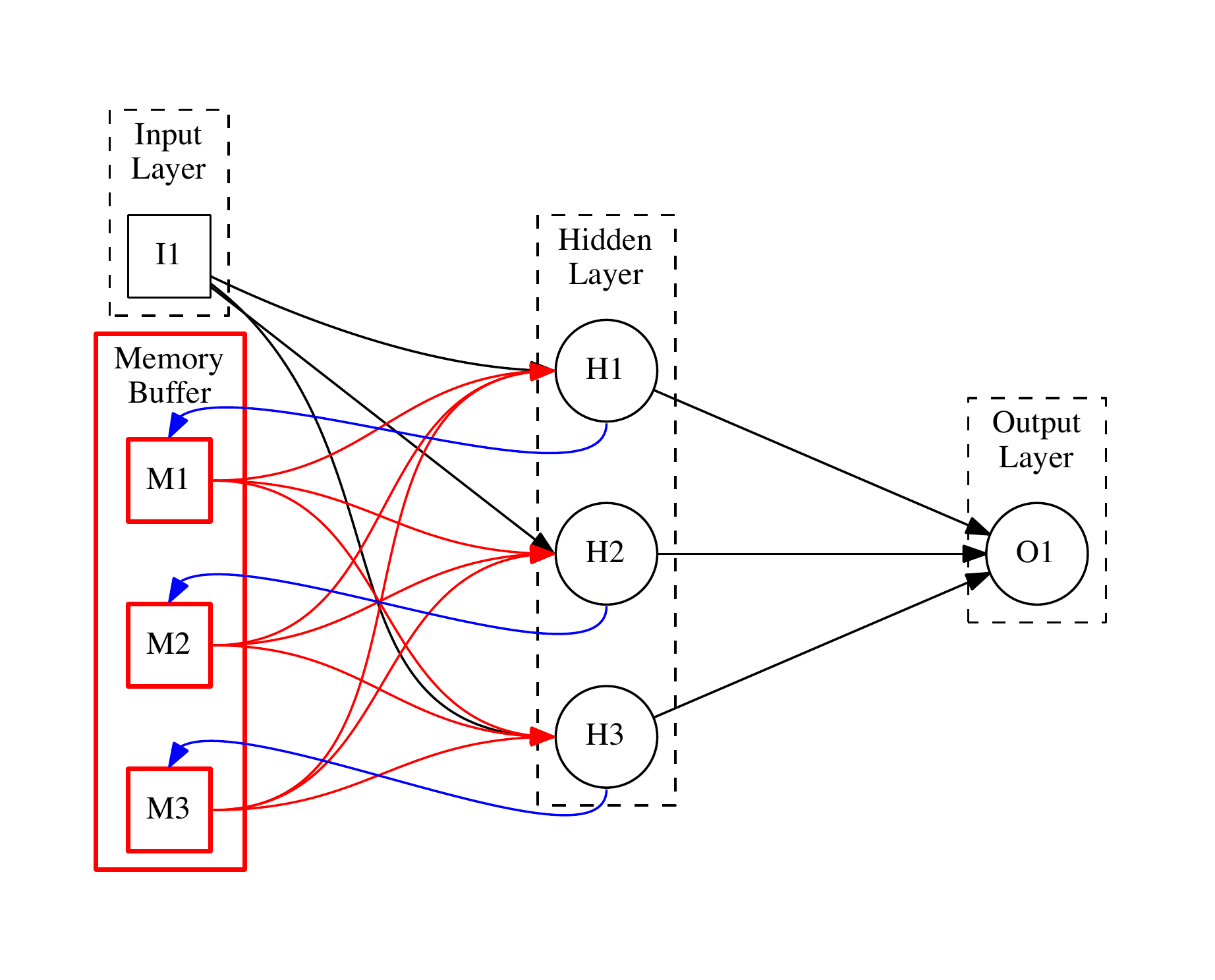}
\caption{The cycle of writing to memory and merging with the next input as the network processes a sequence}
\label{fig:rnn-cycle}
    \end{subfigure}
    \caption{The flow of data between the memory buffer and the hidden layer in a recurrent neural network}
    \label{fig:rnnintro}
\end{figure*}

Recurrent Neural Networks (RNN)\footnote{This introduction to Recurrent Neural Networks is based on \cite{kelleher:2016NMTutorial}.} are an ideal neural network architecture for processing \emph{sequential} data such as language. Generally, RNN models are created by extending a feed-forward neural network that has just one hidden layer 
with a memory buffer, as shown in Figure~\ref{fig:rnn-memorybuffer}. 

RNNs process sequential data one input at a time. In an RNN the outputs of the neurons in the hidden layer of the network for one input are feed back into the network as part the next input. Each time an input from a sequence is presented to the network the output from the hidden units for that input are stored in the memory buffer overwriting whatever was in the memory (Figure~\ref{fig:rnn-writing-to-memory}). At the next time step when the next data point in the sequence is considered, the data stored in the memory buffer is merged with the input for that time step  (Figure~\ref{fig:rnn-reading-from-memory}). Consequently, as the network moves through the sequence there is a recurrent cycle of storing the state of the network and using that state at the next time step (Figure~\ref{fig:rnn-cycle}).

\begin{figure}[htb]
			\centering
        \begin{tikzpicture}[node distance=1cm, auto]  
        \tikzset{
            mynode/.style={rectangle,rounded corners,draw=black, top color=white, bottom color=white,very thick, inner sep=1em, minimum size=3em, text centered, text width=3em},
            myarrow/.style={->, >=latex', shorten >=1pt, thick},
            mylabel/.style={text width=4em, text centered} 
        }  
        \node[mynode] (output) {Output};  
        \node[mynode,  below=of output] (hidden) {Hidden};  
        \node[mynode, below=of hidden] (input) {Input};
        \node[mynode, left=of input] (context) {Memory};
        \draw[myarrow] (hidden.north) --  (output.south);	
        \draw[myarrow] (context.east) --  (hidden.south);
        \draw[myarrow] (hidden.west) --  (context.north);	
        \draw[myarrow] (input.north) --  (hidden.south);
        \end{tikzpicture} 

        (a)
       \bigskip

        \begin{tikzpicture}[node distance=1cm, auto]  
        \tikzset{
            mynode/.style={rectangle,rounded corners,draw=black, top color=white, bottom color=white,very thick, inner sep=1em, minimum size=3em, text centered, text width=3em},
            myarrow/.style={->, >=latex', shorten >=1pt, thick},
            mylabel/.style={text width=4em, text centered} 
        }  
        \node[mynode] (output) {$\mathbf{y}_{t}$};  
        \node[mynode,  below=of output] (hidden) {$\mathbf{h}_{t}$};  
        \node[mynode, below=of hidden] (input) {$\mathbf{x}_{t}$};
        \node[mynode, left=of input] (context) {$\mathbf{h}_{t-1}$};
        \draw[myarrow] (hidden.north) --  (output.south);	
        \draw[myarrow] (context.east) --  (hidden.south);
        \draw[myarrow] (hidden.west) --  (context.north);	
        \draw[myarrow] (input.north) --  (hidden.south);
        \end{tikzpicture} 

        (b)
        \caption{Recurrent Neural Network (RNN)} 
        \label{fig:rnn-abstract}
        \end{figure}
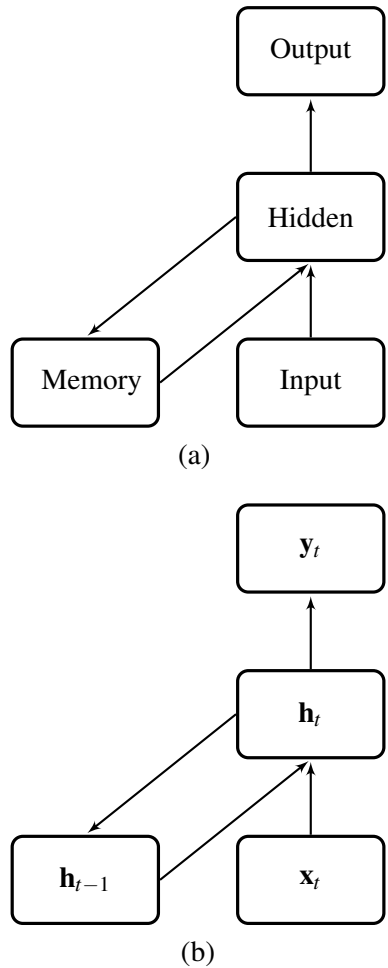

In order to 
simplify the following figures 
we do not draw the individual neurons and connections 
but represent each layer of neurons as a rounded box and show the flow of information between layers with arrows. Also, 
we refer to the input layer as $x_t$, the hidden layer as $h_t$, the output layer as $y_t$, and the memory layer as $h_{t-1}$. Figure~\ref{fig:rnn-abstract}a illustrates the use of this schematic representation of 
layers of neurons and the flow of information through an RNN 
and Figure~\ref{fig:rnn-abstract}b shows the same network using the shorter naming convention. 

\begin{figure}[htb]
	\includegraphics[width=\columnwidth]{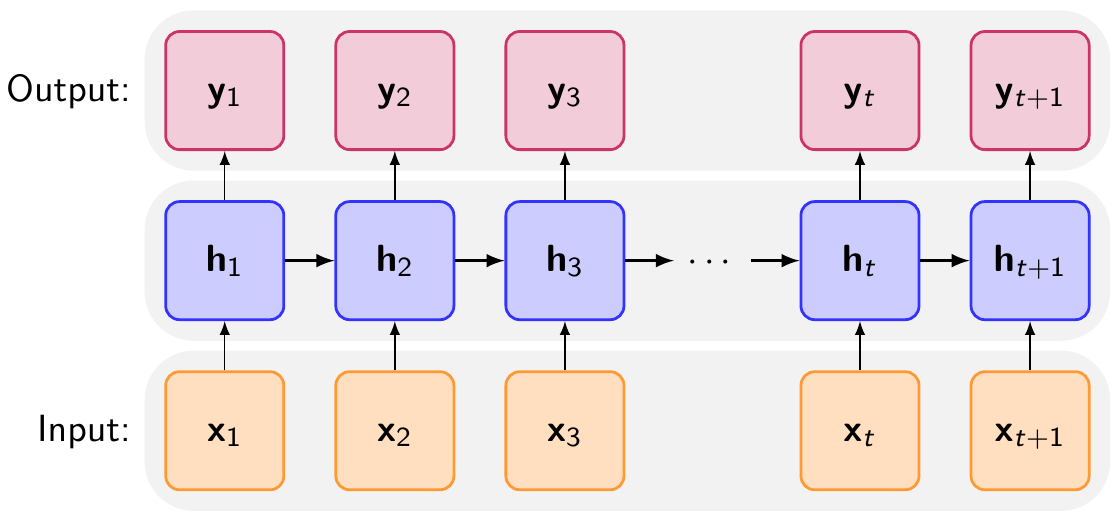} 
	\caption{An RNN unrolled in time}
	\label{fig:rnn-throughtime}
\end{figure}

Figure \ref{fig:rnn-throughtime} demonstrates 
the flow of information through an RNN as it processes a sequence of inputs. An interesting thing to note is that there is a path connecting each $h$ (the hidden layer for each input) to all the previous $h$s. Thus, the hidden layer in an RNN at each point in time is dependent on its past. In other words, the network has a memory so that when it is making a decision at time step $t$ it can remember what it has seen previously. This allows the model to take into account data that depends on previous data, for example in sequences. This is the reason why an RNN is useful for language processing: having a memory of the previous words that have been observed in a sequence of a sentence is predictive of the words that follow them.

\begin{figure}[htb]
		\includegraphics[width=\columnwidth]{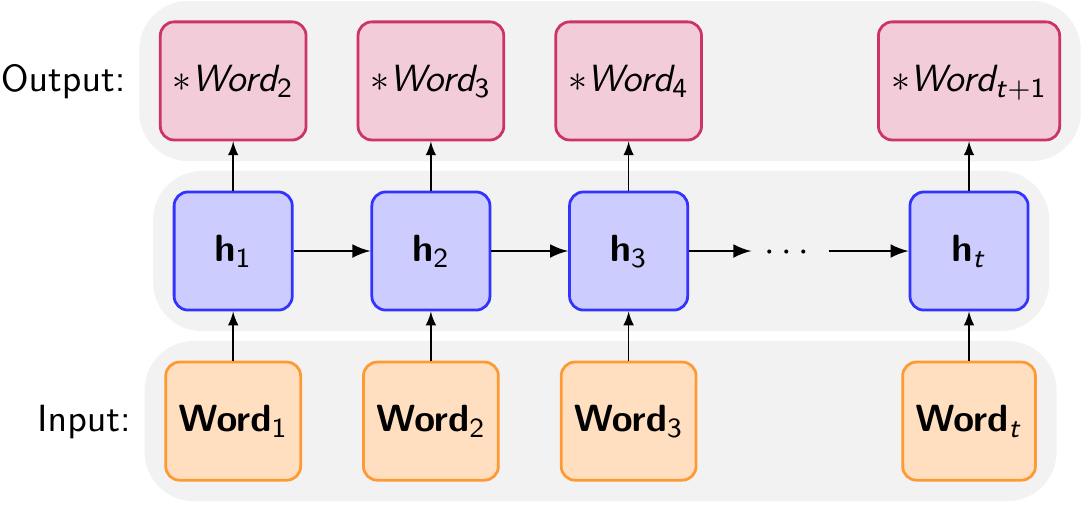} 
		\caption{RNN language model unrolled in time}
		\label{fig:rnn-language-model}
	\end{figure}

A language model is a computational model that takes a sequence of words as input and returns a probability distribution 
from which the probability of each vocabulary word being the next word in the sequence can be predicted. 
An RNN language model can be trained to predict the next word in a sequence. Figure~\ref{fig:rnn-language-model} illustrates how information flows through an RNN language model as it processes a sequence of words and attempts to predict the next word in the sequence after each input. The * indicates the next word as predicted by the system. All going well $*Word_2=Word_2$ but if the system makes a mistake this will not be the case.

\begin{figure}[htb]
		\includegraphics[width=\columnwidth]{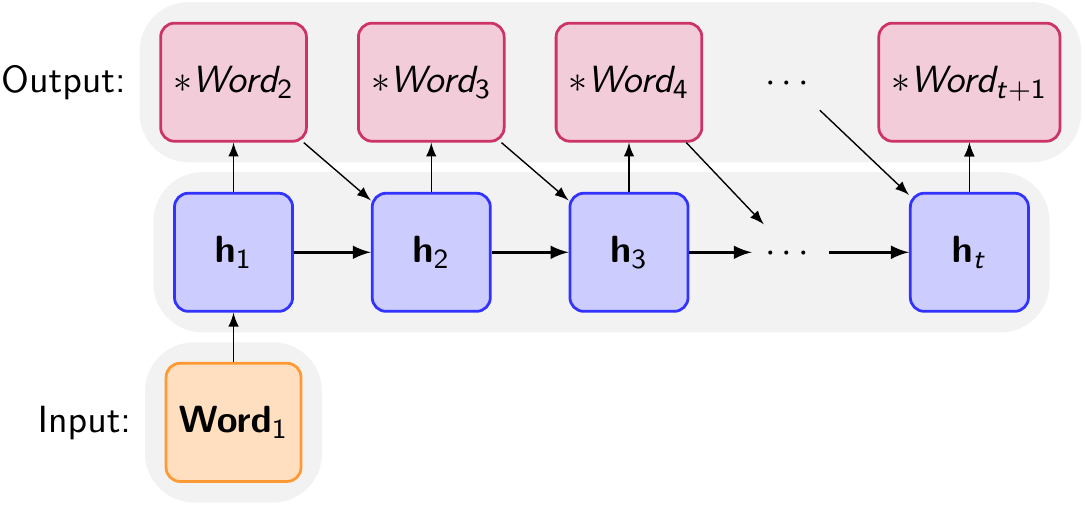} 
		\caption{Using an RNN language model to generate a word sequence}
		\label{fig:rnn-hallucinating}
	\end{figure}

When we have trained a language model we can make it to ``hallucinate'' or generate language by giving it an initial word and then inputting the word that the language model predicts as the most likely next word as the following input word into the model, etc. Figure~\ref{fig:rnn-hallucinating} shows how we can use an RNN language model to generate text by feeding the words the language model predicts back into the model. 

\section{Grounding Spatial Language in Perception}
\label{sec:grounding}

The \emph{symbol grounding problem} is the problem of how the meaning of a symbol can be grounded in anything other than other meaningless symbols. 
\citet{Harnad:1990} argues the symbolic representations must be grounded bottom-up from two forms of non-symbolic sensor representations: \emph{iconic representations} which can be understood as sensory experience of objects and events, and \emph{categorial representations} which are feature detectors that are triggered by invariant features of objects and events within these sensory experiences. Given these two foundational non-symbolic representations, a grounded symbolic system can be built up with the elementary symbols of this system being the symbolic names or labels of the object and event categories that are distinguished within the categorical representations of the agent. Essentially, the meaning of an elementary symbol is the categorisation of sensor grounded experience. 

The description ``meaningless'' in the definition above (originally made by \cite{Harnad:1990}) should be discussed in relation to the work in distributional semantics \cite{Turney:2010aa} which has been used very successfully for computational representation of meaning. The reason why distributional semantic representations work is that word contexts capture indirectly latent situations that co-occuring words are all referring to. Distributional semantic models are built from grounded language (which is therefore not ``meaningless'') it is only that grounded representations are not included in the model. Grounding is expressed indirectly through word co-occurrences.

\cite{roy2005semiotic} extends Harnad's approach by setting out a framework of semiotic schemas that ground symbolic meaning in a causal-predictive cycle of action and perception. Within Roy's framework meaning is ultimately grounded in schemas where each schema is a belief network that connects action, perception, attention, categorisation, inference, and prediction. These schemas can be understood as the interface between the external world (reached through the action and perception components of the schemas) and the agents internal cognitive processes (attention, categorisation, inference, and prediction).

Spatial language is an interesting case study in grounding language in perception because linguistic descriptions of perceived spatial relations between objects are intrinsically about the world and as such should be grounded within an agent's perception of that world \cite{Dobnik:2009dz,Kelleher:2009fk}. The most common form of spatial language discussed in the literature is a \emph{locative expression}. A locative expression is composed of a noun phrase modified by a prepositional phrase that specifies the location of the referent of the noun phrase relative to another object. We will use the word \emph{target object} to refer to the object whose position is being described and the term \emph{landmark object} to refer to the object that the target object's location is described relative to\footnote{The literature on locative expressions uses uses a variety of terms to describe the target and the landmark objects, for a review see \cite{kelleher:2003,Dobnik:2009dz}. Other terms found in the literature for the target object include:  \emph{trajector}, \emph{local object}, and \emph{figure object}. Other terms used to describe the landmark object include: \emph{reference object}, \emph{relatum}, and  \emph{ground}.}, the annotations on the following example locative expression illustrates this terminology:

\begin{center}
$\underbrace{\underbrace{\underbrace{The~big~red~book}_{Target}~\underbrace{on~\underbrace{the~table}_{Landmark}}_{\substack{Prepositional\\Phrase}}}_{Noun~Phrase}}_{Locative~Expression}$
\end{center}



Previous work on spatial language has revealed a range of factors that impinge on the interpretation of locative expressions. An obvious component in the grounding of a spatial description is the scene geometry and the size and shape of region described by the spatial term within that geometry. The concept of a \emph{spatial template} is used to describe these regions and several experiments have revealed how these templates vary across spatial terms and languages, e.g., \cite{logan/sadler:1996,kelleher/costello:2005,Dobnik:2017ac}. 

It has also been shown that the geometry of a spatial template for a given preposition is affected by a number cognitive and contextual factors, including: 
\begin{itemize}[noitemsep]
	\item perceptual attention \cite{regier/carlson:2001,burigo2015visual,kluth2014attentional}, 
	\item the viewer's perspective on the landmark(s) \cite{kelleher/vanGenabith:2006}, 
	\item object occlusion \cite{kelleher2011effect}, 
	\item frame of reference ambiguity and alignment between the landmark object and reference frames \cite{carlson-radvansky/logan:1997,burigo2004reference,kelleher/costello:2005}, 
	\item the location of other \emph{distractor objects} in the scene \cite{kelleher2005context,costellokelleher:06,kelleher/etal:2006},
        \item and the richness of the perceptual context \cite{Dobnik:2017ac}.
\end{itemize}

The last point is related to the fact that factors affecting semantics of spatial descriptions go beyond scene geometry and include the functional relationships between the target and the landmark \cite{coventry:1998,CoventryEtAl:2001,Coventry/Garrod:2004} and force dynamics within the scene \cite{Coventryetal:2005,Sjoo:2011aa}. These functional relations can be captured as meanings induced from word distributions \cite{Dobnik:2013aa,dobnik2014exploration}. Another important factor of (projective) spatial descriptions is their contextual underspecification in terms of the assigned frame of reference which is coordinated through dialogue interaction between conversational participants \cite{Dobnik:2015aa}. It is therefore based on their coordinated intentions in their interaction.

The research in spatial language semantics highlights its multifaceted nature. Spatial language draws upon (i) geometric concepts, (ii) world knowledge (i.e., an understanding of functional relationships and force dynamics), and (iii) perceptual and discourse cues. Thus in order for a computational system to adequately model spatial language semantics it should accommodate all or most of these factors.


\section{Spatial Language in DL}
\label{sec:spatial-dl}

The question that this paper addresses is whether deep learning image captioning architectures as currently constituted are capable of grounding spatial language within the images they are captioning. The outputs of these systems are impressive and often include spatial descriptions. Figure~\ref{fig:imagecaptioningbird} (based on an example from \cite{xu2015show}) provides an indicative example of the performance of these systems. The generated caption in this case is accurate and what is particularly interesting is that it includes a spatial description: \emph{over water}. Indeed, the vast majority of generated captions listed in \cite{xu2015show} include spatial descriptions, some of which include:\footnote{The emphasis on the spatial descriptions were added here.}

\begin{itemize}[noitemsep]
	\item ``A woman is throwing a frisbee \emph{in a park}''
	\item ``A dog is standing \emph{on a hardwood floor}''
	\item ``A group of people sitting \emph{on a boat in the water}''.
\end{itemize}

The fact that these example captions include spatial descriptions and that the captions are often correct descriptions of the input image begs the question of whether image captioning systems are actually learning to ground the semantics of these spatial terms in the images. The nature of neural network systems makes it difficult to directly analyse what a system is learning, however there are a number of reasons why it would be surprising to find that these systems were grounding spatial language. First, recall from the review of grounding in Section~\ref{sec:grounding} that spatial language draws on a variety of information, including:

\begin{itemize}[noitemsep]
	\item scene geometry,
	\item perceptual cues such as object occlusion,
	\item world knowledge including functional relationships and force dynamics,
        \item and coordinated intentions of interacting agents.
\end{itemize}

Considering only scene geometry, these image captioning systems use CNNs to encode the representation of the input image. Recall from Section~\ref{sec:cnn} that CNNs discard locational information through the (down-sampling) pooling mechanism and that such down-sampling may be applied several times within a CNN pipeline. Although it is possible that the encoding generated by a CNN may capture rough relative positions of objects within a scene, it is likely that this encoding is too rough to accommodate the level of sensitivity of spatial descriptions to location that experimental studies of spatial language have found to be relevant (cf. the changes in acceptability ratings in \cite{logan/sadler:1996,kelleher/costello:2005,Dobnik:2017ac} as a target object moved position relative to the landmark). The architecture of CNNs also points to the fact that these systems are unlikely to be modelling perceptual cues. CNNs essentially work by identifying what an object is through a hierarchical merging of local visual features that are predictive of the object type. These local visual features are likely to be features that are parts of the object and therefore frequently co-occur with the object label. Consequently, CNNs are unlikely to learn to identify an object type via context and viewpoint dependent cues such as occlusion. Finally, neither a CNN nor an RNN as currently used in the image description tasks provide mechanisms to learn force-dynamics or functional relationships between objects (cf. \cite{Coventryetal:2005,battaglia2013simulation}) nor do they take into account agent interaction. Viewed in this light the current image captioning systems appear to be missing most of the key factors necessary to ground spatial language. And, yet they do appear to generate reasonably accurate captions that include spatial descriptions.

There are a number of factors that may be contributing to this apparent ability. First, an inspection of the spatial descriptions used in the generated captions reveals that they tend to include \emph{topological} rather than \emph{projective} spatial prepositions (e.g., \emph{on} and \emph{in} rather than \emph{to the left of} and \emph{above}): \emph{in a forest}, \emph{in a park}, \emph{in a field}, \emph{in the field with trees}, \emph{in the background},  \emph{on a bed}, \emph{on a road}, \emph{on a skateboard}, \emph{at a table}. These spatial descriptions are more underspecified with regard to the location of the target object relative to the landmark object than projective descriptions which also require grounding of direction within a frame of reference. Topological descriptions are semantically adequate already if the target is just proximal to the landmark and hence it is more likely that a caption will be acceptable. Furthermore, it is frequently the case that given a particular label for a landmark it is possible to guess the appropriate preposition irrespective of the image and/or the target object type or location. Essentially the task posed to these systems is to fill the blanks with one of \emph{at}, \emph{on}, \emph{in}: 

\begin{itemize}[noitemsep]
	\item \emph{TARGET} \emph{\rule{.2\columnwidth}{0.4pt} a field},
	\item \emph{TARGET} \emph{\rule{.2\columnwidth}{0.4pt} the background},  
	\item \emph{TARGET} \emph{\rule{.2\columnwidth}{0.4pt} a road}, 
	\item \emph{TARGET} \emph{\rule{.2\columnwidth}{0.4pt} a table}. 
\end{itemize}

Although the system may get some of the blanks wrong it is likely to get many of them right. This is because the system can use distributional knowledge of words which captures some grounding indirectly as discussed in Section~\ref{sec:grounding}. Indeed, recent research has shown that co-occurrence of nouns with a preposition within a corpus of spatial descriptions can reveal functional relations between objects referred to by the nouns \cite{Dobnik:2013aa,dobnik2014exploration}. Word co-occurrence is thus highly predictive of the correct preposition. Consequently, language models trained on image description corpora indirectly model partially grounded functional relations, at least within the scope of the co-occurrence likelihood of prepositions and nouns. 

The implication of this is that current image captioning systems do not ground spatial descriptions in the images they take as input. Instead, the apparent ability of these systems to frequently correctly use spatial prepositions to describe spatial relations within the image is the result of the RNN language model learning to predict the most likely preposition to be used given the target and landmark nouns where these nouns are predicted from the image by the CNN. 

There is a negative and a positive side to this conclusion. Let's start with the negative side. The distinction between cognitive representations of \emph{what} something is versus \emph{where} something is has a long tradition in spatial language and spatial cognition research \cite{landauJackendoff:1993}. These image captioning systems would appear to be learning representations that allow them to ground the semantics of \emph{what}. But they are not learning representations that enable them to ground the semantics of \emph{where}. Instead, they rely on the RNN language model to make good guesses of the appropriate spatial terms to use based on word distributions. The latter point introduces the positive side. It is surprising how much and how robustly semantic information can be captured by distributional language models. Of course, language models cannot capture the geometric relations between objects, for example they are not able to distinguish successfully the difference in semantics between \emph{the chair is to the left of the table} and \emph{the chair is to the right of the table} as \emph{left} and \emph{right} would occur in exactly the same word contexts. However, as we argued in Section~\ref{sec:grounding} spatial language is not only spatial but also affected by other sources of knowledge that leave an imprint in the word distributions which capture relations between  higher-level categorical representations built upon the elementary grounded symbols \cite{Harnad:1990}. It follows that some categorical representations will be closer to and therefore more grounded in elementary symbols, something that has been shown for spatial language \cite{coventry:1998,CoventryEtAl:2001,Coventry/Garrod:2004,Dobnik:2013aa,dobnik2014exploration}. In conclusion, it follows that successful computational models of spatial language require both kinds of knowledge.

\section{Conclusions}
\label{sec:conclusions}

In this paper we examined the current architecture for generating image captions with deep learning and argued that in its present setup they fail to ground the meaning of spatial descriptions in the image but nonetheless achieve a good performance in generating spatial language which is surprising given the constraints of the architecture that they are working with. The information that they are using to generate spatial descriptions is not spatial but distributional, based on word co-occurrence in a sequence as captured by a language model. While such information is required to successfully predict spatial language, it is not sufficient. We see at least two useful areas of future work. On one hand, it should be possible to extend the deep learning configurations for image description to take into account and specialise to learn geometric representations of objects, just as the current deep learning configurations are specialised to learn visual features that are indicative of objects. The work on modularity of neural networks such as \cite{Andreas:2016aa,Johnson:2017aa} may be relevant in this respect. On the other hand, we want to study how much information can be squeezed out of language models to successfully model spatial language and what kind of language models can be built to do so.



\section*{Acknowledgements}

The research of Kelleher was supported by the ADAPT Research Centre. The ADAPT Centre for Digital Content Technology is funded under the SFI Research Centres Programme (Grant 13/RC/2106) and is co-funded under the European Regional Development Funds.

The research of Dobnik was supported by a grant from the Swedish Research Council (VR project 2014-39) for the establishment of the Centre for Linguistic Theory and Studies in Probability (CLASP) at Department of Philosophy, Linguistics and Theory of Science (FLoV), University of Gothenburg.

\bibliography{references}
\bibliographystyle{acl_natbib}

\end{document}